\documentclass[runningheads]{llncs}
\usepackage{subfig}
\usepackage{hyperref}
\usepackage{url}

\usepackage{amsmath,epsfig}
\usepackage{amscd}
\usepackage{subcaption}
\usepackage{color, colortbl}
\usepackage{arydshln}
\usepackage{nicematrix,tikz}
\usepackage{algorithm} 
\usepackage{algpseudocode}

\newcommand{\comment}[1]{}

\begin{document}

\title{ A Conformal Prediction Score that  is Robust to Label Noise}
\author{
  Coby Penso  \qquad Jacob Goldberger \\ 
    \texttt{coby.penso24@gmail.com} \qquad \texttt{jacob.goldberger@biu.ac.il} \\
}
\institute{Bar-Ilan University, Israel}

\maketitle

\begin{abstract}
 Conformal Prediction (CP) quantifies network uncertainty by building a small prediction set with a pre-defined probability that the correct class is within this set. In this study we tackle the problem of CP calibration based on a validation set with noisy labels. We introduce a conformal score that is robust to label noise. The noise-free conformal score is estimated using the noisy labeled data and the noise level.
 In the test phase the
 noise-free score is used to form the prediction set.
 We applied the proposed algorithm to several standard medical imaging classification datasets. We show that our method outperforms current methods by a large margin, in terms of the average size of the prediction set, while maintaining the required coverage.

\end{abstract}

\section{Introduction}

In medical imaging classification tasks, it is crucial not only to report the most likely diagnosis but equally or even more important to eliminate other possibilities. Reporting decisions in terms of a prediction set of class candidates is thus a natural approach for clinical applications. In this procedure, the network eliminates most classes, and thus presents the physician with a small subset of options for further investigation. In terms of confidence,  the prediction set must encompass the true diagnosis with a high probability, such as 90\%.

Conformal Prediction (CP) 
\cite{vovk2005conformal,angelopoulos2023conformal} is a general non-parametric prediction-set calibration method. 
Given a required confidence level, it aims to construct a small prediction set with the guarantee that the probability of the correct class being within this set meets or exceeds this requirement."
Recently, CP has become a major calibration tool for neural network systems in various applications including medical imaging \cite{lu2022fair,lu2022improving,olsson2022estimating}. CP is not a specific algorithm but rather a general framework where selecting a specific conformal score defines how the prediction set is constructed. 

Training a neural network requires massive amounts of carefully labeled data to succeed, but acquiring such data is expensive and time-consuming. Medical imaging datasets often contain noisy labels due to ambiguous images that can confuse clinical experts.  Physicians may disagree on the diagnosis of the same medical image, resulting in variability in the ground truth label. Therefore, addressing annotation noise is a crucial topic in medical image analysis. Numerous studies have examined the problem of training networks that are resilient to label noise \cite{9729424,Xue2022}. 
The collected annotated dataset is randomly split into training and validation sets. 
Therefore, the existence of noisy labels not only affects the training procedure but also the CP prediction-set calibration step.
While significant efforts have  been  devoted to the problem of  noise-robust network training , the challenge of calibrating the
models has only recently begun to receive  attention \cite{einbinder2022conformal}.

In this study, we address the challenge of applying CP to medical imaging classification networks using a validation set with noisy labels. Network calibration methods are known to be highly susceptible to label noise. Einberder et al. \cite{einbinder2022conformal} suggested  ignoring  label noise and applying the standard CP algorithm on the noisy labeled validation set. This strategy results in  large prediction sets. Here, we introduce a CP algorithm  based on a new conformal score that is robust to label noise.  We applied the  algorithm to several standard medical imaging classification datasets. The experimental results indicated that our method outperformed state-of-the-art methods that handle label noise by a significant margin.

\section{Conformal Prediction and Label Noise}

Consider a network that classifies an input $x$ into $k$ pre-defined classes.
Given a specified coverage $1\!-\!\alpha$ we want to find a small prediction set (i.e. a subset of the classes) that includes the correct class with a probability of at  least $1\!-\!\alpha$.
 A simple approach to achieving
this goal is to include classes from the highest to the lowest probability until their sum just exceeds the threshold of $1\!-\!\alpha$.
While the network output has the mathematical
form of a distribution, this does not necessarily imply that it represents the true class distribution. The network is usually
not calibrated and tends to be over-optimistic [5].

The CP algorithm builds 
a prediction set with a coverage guarantee.
 A conformal score $S(x,y)$ measures the network uncertainty  between $x$ and its true label $y$ (larger scores encode worse agreement). The Homogeneous Prediction Sets (HPS) score \cite{vovk2005conformal} is 
$S_{\scriptscriptstyle \textrm HPS}(x,y)=1-p(y|x;\theta)$, s.t. $\theta$ is the network parameter set.
Another  widely used score is the Adaptive Prediction  Score (APS) \cite{romano2020classification}:
\begin{equation}
 S_{\scriptscriptstyle APS}(x,y) =  \sum_{\{i|p_i \ge p_{y}\}} p_i,
   \label{aps_score}
  \end{equation}
such that $p_i= p(y=i|x;\theta)$ and $p_y$ is the probability of the true label.
The RAPS score \cite{angelopoulos2020uncertainty} is a variant of APS which, in the case of a large number of classes, encourages small prediction sets. It is defined as follows:
\begin{equation}
 S_{\scriptscriptstyle RAPS}(x,y) =  \sum_{\{i|p_i \ge p_{y}\}} p_i + a \cdot \max(0,(nc - b))
      \label{raps_score}
   \end{equation}
s.t. $nc=|\{i|p_i \ge p_{y}\}|$ and  $a,b$ are parameters that need to be tuned.

The CP calibration procedure operates on a given labeled validation set $(x_1,y_1),...,(x_n,y_n)$.
Let $q$ be the $(1\!-\!\alpha)$ quantile of $S(x_1,y_1),...,S(x_n,y_n)$. The prediction set of a new test point $x$ is defined as $C(x)=\{y| S(x,y) \le q\}$. 
The general CP theory guarantees that $1\!-\!\alpha \le p( y\in C(x)) \le 1\!-\!\alpha + \frac{1}{n+1}$,
where $y$ is the unknown true label \cite{vovk2005conformal}. Note that this is a marginal
probability over all possible test points and coverage may be worse or better for some
cases. It can be proved that conditional coverage is, in general, impossible \cite{foygel2021limits}.

We can also define a randomized version of a conformal score. For example in the case of  APS
we define: \begin{equation}
 S_{\scriptscriptstyle rand-APS}(x,y,u) =  \sum_{\{i|p_i > p_{y}\}} p_i + u \cdot p_y,\hspace{1cm} u\sim U[0,1].
  \label{radaps_score}
  \end{equation}
The CP threshold $q$ is  the $(1\!-\!\alpha)$ quantile of
$S(x_1,y_1,u_1),...,S(x_n,y_n,u_n)$ and the test time (random) prediction set is $C(x)=\{y|S(x,y,u)\le q\}$. In the random case we still have a coverage guarantee  (marginalized overall test points $x$  and samplings  $u$ from the uniform distribution). The random version tends to produce prediction sets with smaller sizes \cite{angelopoulos2023conformal}.

Suppose we have a validation set with potentially inaccurate labels.
Let $y_1,...,y_n$ be the correct labels, and let $\tilde{y}_1,...,\tilde{y}_n$ be the corresponding observed corrupted labels.
We assume that the label noise follows a uniform distribution, where with a probability of $\epsilon$,  the correct label is replaced by a randomly selected label uniformly from all the  $k$ classes:
 \begin{equation}
 p(\tilde{y}=j| y=i) = (1\!-\!\epsilon)1_{\{j=i\}} + \epsilon\frac{1}{k}.
 \label{randomflip}
 \end{equation}
The noise is applied to each sample
independently.  This noise model is commonly referred to as the random flip noise.
The challenge is to successfully implement the CP algorithm using a validation set with noisy labels.

Einberder et al. \cite{einbinder2022conformal} suggested  applying the standard CP algorithm to the  noisy labeled validation set. We denote their approach as Noisy-CP.  The general CP theory guarantees that $1\!-\!\alpha \le p( \tilde{y}\in C_{noisy-cp}(x))$.   This guarantee is for the noisy labeled data. They showed that if the classifier $p(y|x;\theta)$ ranks the classes identically to the true distribution $p(y|x)$, we also have a coverage guarantee for the noise-free test data. The CP threshold $q_{noise}$ obtained by the Noisy-CP procedure is usually much higher than the threshold $q$ obtained by using noise-free data. This results in larger prediction sets. Our goal is to find a noise-robust calibration score with the required  $1\!-\!\alpha$  coverage and with a small prediction-set size.

\section{A Noise-Robust Conformal Prediction Score}
\label{sec:method}
Let $(x_1,\tilde{y}_1),...,(x_n,\tilde{y}_n)$ be a noisy validation set where the labels were corrupted by uniform noise with a noise level $\epsilon$.
 We aim to find a noise-robust conformal score that can be applied to the noisy labeled data. 
 Since $y_t$ is not observed, we cannot directly compute the score $S(x_t,y_t)$. Instead, we can estimate it using its noisy version $\tilde{y}_t$:
\begin{equation}
\mathbf{E} ( S(x_t,y_t)| \tilde{y}_t) = \sum_{i=1}^k p(y_t=i|\tilde{y}_t) S(x_t,i).
\label{est}
\end{equation}
Assuming a non-informative uniform  prior on the correct label $y_t$, i.e., $p(y_t=i)=1/k$), we obtain:
\begin{equation}
  p(y_t=i|\tilde{y}_t)  = (1\!-\!\epsilon) 1_{\{\tilde{y}_t=i\}}+ \frac{\epsilon}{k}.
  \label{ytnoise}
  \end{equation}
Substituting  (\ref{ytnoise}) in  (\ref{est}), yields an estimate 
$ \hat{S}(x_t,\tilde{y}_t,\epsilon)$ of the noise-free score:
\begin{equation}
\hat{S}(x_t,\tilde{y}_t,\epsilon) = \mathbf{E} ( S(x_t,y_t)| \tilde{y}_t) = (1\!-\!\epsilon) S ( x_t,\tilde{y}_t)  + \epsilon S(x_t)
\label{avg_score}
\end{equation}
s.t. $S(x_t) =   \frac{1}{k} \sum_{i=1}^k S(x_t,i)$.
Note that to obtain the score estimation $\hat{S}(x_t,\tilde{y}_t,\epsilon)$,  we need to either know the noise level $\epsilon$ or estimate it from the noisy-label data. We elaborate further on this issue in the next section.

\renewcommand{\arraystretch}{1.5}
\begin{table}[t]
\centering
\caption{Conformal Prediction methods for validation sets with noisy labels. Given an image $x$, $y$ is the true label, $\hat{y}$ is its noisy version and $\epsilon$ is the noise level.  $S$ is a conformal score and $\hat{S}$ is its noise robust variant.}
\begin{tabular}{l|c|c|c|c}
\hline
Stage & CP (Oracle) & Noisy-CP \cite{einbinder2022conformal} & NRES-CP & NR-CP  
\\  \hline
Learning phase& $S(x,y) \xrightarrow{} q$ & $S(x,\tilde{y}) \xrightarrow{} q_{noise}$ & $\hat{S}(x,\tilde{y},\epsilon) \xrightarrow{} q_{\epsilon}$ & $\hat{S}(x,\tilde{y},\epsilon) \xrightarrow{} q_{\epsilon}$  \\  \vspace{1mm} \hspace{-2mm}
Inference phase & $\{y|S(x,y) \le q\} $ & $\{y|S(x,y)\! \le \!q_{noise}\} $ & $\{y|\hat{S}(x,y,\epsilon) \le q_{\epsilon}\}$ & $\{y|S(x,y) \le q_{\epsilon}\} $ \\  \hline
\end{tabular}
\label{noisy_methods}
\end{table}
\renewcommand{\arraystretch}{1.0}

We next apply the CP algorithm on the estimated conformal scores and set $q_{\epsilon}$ to be the $(1\!-\!\alpha)$ quantile of $\hat{S}(x_1,\tilde{y}_1,\epsilon),...,\hat{S}(x_n,\tilde{y}_n,\epsilon)$.
According to the general CP theory, the prediction set of a given test sample $x$ is:
  \begin{equation}
  \hat{C}_{\epsilon}(x)=\{ y \,|\, \hat{S}(x,{y},\epsilon) \le q_{\epsilon}\}  =
 \{ y \, | \, S (x, y)  \le \frac{q_{\epsilon}-\epsilon S(x)}{1\!-\!\epsilon}\}.
 \label{nrescp}
\end{equation}
Let $x$ be a test point and $y$ and $\tilde{y}$ be its true label and noisy label respectively.
  The general CP theory guarantees that $1\!-\!\alpha \le p( \tilde{y}\in \hat{C}_{\epsilon}(x))$. This guarantee, however, is for the noisy labeled data. Assume that for every $x$ the order of the network class predictions  $p(y=i|x;\theta)$ coincides with the order of the true probabilities $p(y=i|x)$. In that case, the same argument that appears in  \cite{einbinder2022conformal} for Noisy-CP, implies a coverage guarantee in the noise-free case.
However, the prediction set obtained by the estimated score  (\ref{nrescp}), is usually still too large.

In the case of noisy labels, during the CP learning phase, we need to  estimate the score of the correct class. However,  to form the prediction set at test time we need to compute the scores of all the possible classes. Hence, 
in a way similar to the Noisy-CP algorithm,
\begin{algorithm}[t]
\caption{Noise-Robust Conformal Prediction (NR-CP)}
       \begin{algorithmic}[1]
      \State Input: A conformal score $S(x,y)$, a coverage level $1\!-\!\alpha$ and a validation set $(x_1,y_1),...,(x_n,y_n)$,
      s.t. the labels are corrupted by uniform noise with parameter~$\epsilon$.  
    \State Compute the estimated scores:
    $$  s_t = \hat{S}(x_t,y_t,\epsilon) = (1-\epsilon) S ( x_t,y_t)  + \frac{\epsilon}{k} \sum_{i=1}^k {S}(x_t,i), \hspace{0.5cm} t=1,...,n$$  
    \State Set $q$ to be the 
    $\lceil(n+1)(1\!-\!\alpha)/n\rceil$ quantile of $s_1,...,s_n$.
        \vspace{1mm}
    \State The prediction set of a test sample $x$
         is $C(x)=\{y\,|\,S(x,y)<q\}$. 
   \end{algorithmic}
   \label{alg1}
       \end{algorithm}
we can thus construct the prediction set of a test sample $x$ using the exact score $S(x,y)$  instead of the estimated score $\hat{S}(x,\tilde{y},\epsilon)$:
\begin{equation}
C_{\epsilon}(x)=\{ y \,| \,S(x,y) \le q_{\epsilon}\}.
\label{nrcp}
\end{equation}
 We denote the algorithm variants based on Eqs. (\ref{nrescp}) and (\ref{nrcp}) as the  Noise Robust Estimated Score CP  (NRES-CP) and the Noise Robust  CP (NR-CP) respectively. 
 The only difference between them lies in how the test-time prediction set is formed.
 The various noisy label CP methods discussed above are summarized in Table~\ref{noisy_methods}.
We empirically show below that NR-CP satisfies the coverage requirement and yields an average size that is much smaller than the one obtained by NRES-CP. The NR-CP is summarized in Algorithm Box 1.

We next analyze the proposed noise robust conformal score.
It is easy to verify that the prediction set obtained by NR-CP is smaller than the one obtained by NRES-CP, i.e.,  $C_{\epsilon}(x)  \subset \hat{C}_{\epsilon}(x)$ if and only if  $S(x) \le q_{\epsilon}$.
In the case of HPS, $S(x)=\frac{1}{k}\sum_i (1-p(y=i|x))=(k-1)/k$ and therefore $\hat{S}(x,\tilde{y},\epsilon) = (1-\epsilon)  S(x,\tilde{y})+ \epsilon\frac{k-1}{k}$.
This implies that  $q_{\epsilon}  = (1-\epsilon)  q_{noise} + \epsilon\frac{k-1}{k}$ 
such that $q_{noise}$ and $q_{\epsilon}$ are the thresholds computed by Noisy-CP \cite{einbinder2022conformal} and NRES-CP (\ref{nrescp}) respectively. It is easy to verify that Noisy-CP  and NRES-CP yield the same prediction set, i.e. 
$\{y|S(x,y) \le q_{noise}\}   = \{y|\hat{S}(x,y,\epsilon) \le q_{\epsilon}\}$.  

  In the case of the APS conformal score, 
  it is easy to see that:  $
  (\hat{p}+1)/2 \le  S(x) \le (\hat{p} + k-1)/k$,
 s.t.  $ \hat{p}=  \max_i p(y=i|x;\theta)$ is the network confidence on its single-class prediction.
      Hence,  when the prediction sets obtained by NR-CP are smaller than those obtained by NRES-CP, the prediction confidence satisfies $(\hat{p}+1)/2 \le S(x) \le  q_{\epsilon}$. In this case the network is less confident       ($\hat{p} \le 2 q_{\epsilon}-1$)
      and thus the size prediction set is expected to be larger.
\begin{table}[th]
\centering
\caption{ APS Caibration results for $1\!-\!\alpha$ = 0.9.  We report the mean and the std over 5000 different splits.}
\begin{tabular}{l|lcc|c}
\hline
Dataset      &  CP Method & size $\downarrow$ & $q$ (\%)    & coverage (\%)           \\ \hline
TissueMNIST \cite{medmnistv2} &  CP & 3.47 $\pm$ 0.02 & 98.59 $\pm$ 0.05  & 90.00 $\pm$ 0.22 \\ 
      (8 classes) &  Noisy-CP & 5.14 $\pm$ 0.03 & 99.89 $\pm$ 0.01  & 97.64 $\pm$ 0.11 \\ 
                  &  NRES-CP & 6.16 $\pm$ 0.03 & 99.16 $\pm$ 0.03   & 93.85 $\pm$ 0.15 \\ 
                  &  NR-CP & \textbf{3.87 $\pm$ 0.02} & 99.16 $\pm$ 0.03  & 92.63 $\pm$ 0.16 \\ \hline 
PathMNIST   \cite{medmnistv2} &  CP & 4.27 $\pm$ 0.09 & 97.49 $\pm$ 0.31  & 90.03 $\pm$ 0.71 \\ 
      (9 classes) &  Noisy-CP & 6.21 $\pm$ 0.11 & 99.38 $\pm$ 0.15  & 96.06 $\pm$ 0.39 \\ 
                  &  NRES-CP & 6.10 $\pm$ 0.15 & 98.60 $\pm$ 0.17  & 94.90 $\pm$ 0.43 \\ 
                  &  NR-CP & \textbf{5.16 $\pm$ 0.09} & 98.60 $\pm$ 0.17  & 93.64 $\pm$ 0.41 \\ \hline 
HAM10000  \cite{tschandl2018ham10000}   &  CP & 3.95 $\pm$ 0.21 & 97.55 $\pm$ 0.39  & 90.20 $\pm$ 1.78 \\ 
      (7 classes) &  Noisy-CP & 5.10 $\pm$ 0.19 & 99.05 $\pm$ 0.21  & 95.41 $\pm$ 1.07 \\ 
                  &  NRES-CP & 5.12 $\pm$ 0.18 & 98.18 $\pm$ 0.17   & 93.92 $\pm$ 1.23 \\
                  &  NR-CP & \textbf{4.37 $\pm$ 0.13} & 98.18 $\pm$ 0.17  & 92.36 $\pm$ 1.10 \\ \hline 
OrganSMNIST \cite{medmnistv2} &  CP & 4.07 $\pm$ 0.07 & 99.6 $\pm$ 0.04  & 90.02 $\pm$ 0.57 \\ 
      (11 classes) &  Noisy-CP & 6.67 $\pm$ 0.12 & 99.96 $\pm$ 0.01  & 98.40 $\pm$ 0.20 \\ 
                  &  NRES-CP & 6.52 $\pm$ 0.13 & 99.84 $\pm$ 0.02  & 94.87 $\pm$ 0.36 \\ 
                  &  NR-CP & \textbf{5.20 $\pm$ 0.07} & 99.84 $\pm$ 0.02  & 95.19 $\pm$ 0.34 \\ \hline 
\end{tabular}

\label{table_aps}
\end{table}

\section{Experiments and Results}
In this section, we evaluate the capabilities of our NR-CP algorithm on various medical imaging datasets. We share our code for reproducibility\footnote{\url{https://github.com/cobypenso/Noise-Robust-Conformal-Prediction}}.

 \textbf{Compared methods.} Our method takes an existing conformal score $S$ and creates a variant of it $\hat{S}$ that is robust to label noise.  We  implemented the three most popular conformal prediction scores, namely APS 
\cite{romano2020classification}, RAPS 
\cite{angelopoulos2020uncertainty}
 and HPS \cite{vovk2005conformal}. For each score $S$, we compared four CP  methods: (1) CP (Oracle) -  using a validation set with clean labels,  (2) Noisy-CP -  applying a standard CP on noisy labels without any modifications \cite{einbinder2022conformal}, (3) NRES-CP and (4) NR-CP. 
 Table~1 summarizes the differences between methods.   

{\bf Evaluation Measures}. The evaluation was done on a noise-free test set. We evaluated each method   based on the average length of the prediction sets (where a small value means high efficiency) and the fraction of test samples for which the prediction sets contain the ground-truth labels. These two evaluation metrics can
be formally defined as:
$$ \textrm{size} = \frac{1}{n} \sum_i | C(x_i) |,  \hspace{1cm} 
 \textrm{coverage} = \frac{1}{n} \sum_i 
{\bf 1}(y_i \in C(x_i))$$
such that  $n$ is the size of the test set.

 \textbf{Datasets.}
We present results on several publicly available medical imaging classification datasets.
 \textbf{TissuMNIST}
 \cite{medmnistv1,medmnistv2}:
It contains 236,386 human kidney cortex cells,  organized into 8 categories. Each gray-scale
image is $32 \times 32 \times 7$ pixels. The  2D projections were obtained by taking the maximum pixel value along the axial-axis of each pixel, and were resized into $28 \times 28$ gray-scale images \cite{woloshuk2021situ}. 
 \textbf{HAM10000} \cite{tschandl2018ham10000}: This dataset contains 10,015 dermatoscopic images of size $800 \times 600$. Cases include a representative collection of 7 diagnostic categories in the realm of pigmented lesions. We used a train/validation/test split of 8,013/1,001/1,001 images.
 \textbf{PathMNIST} \cite{medmnistv2}: A dataset that contains 97,176 images of colon pathology with nine classes. The images'  size is $28 \times 28$. Here, we used a train/validation/test split of 89,996/3,590/3,590 images.
 \textbf{OrganSMNIST} \cite{medmnistv2}: A dataset that contains 25,221 images of abdominal CT with eleven classes. The images are  is $28 \times 28$ in size. Here, we used a train/validation/test split of 13,940/2,452/8,829 images.

\textbf{Implementation details.} Each task was trained by fine-tuning on a pre-trained ResNet-18 \cite{he2016deep} network. The models were taken from the PyTorch site\footnote{
\url{https://pytorch.org/vision/stable/models.html}}. This network architecture was selected because of its widespread use in classification problems.
The last fully connected layer output size was adjusted to fit the corresponding number of classes for each dataset. For the HAM10000 models we fine-tuned using cross-entropy
loss and the Adam optimizer, whereas for PathMNIST, TissueMNIST, and OrganSMNIST we used publicly available 
checkpoints\footnote{\url{https://github.com/MedMNIST/MedMNIST}}.
For each dataset, we combined the validation and test sets and then constructed 5000 different splits where 50\% was used for the calibration phase and 50\% used for testing.

\begin{table}[t]
\centering
\caption{ APS and RAPS (randomized versions) calibration results for $1\!-\!\alpha$ = 0.9.  We report the mean and the std over 5000 different splits.}
\label{raps}
   \scalebox{.95}{ 
\begin{tabular}{ll|cc|cc}
& & \multicolumn{2}{c|}{rand-APS}  & \multicolumn{2}{c}{rand-RAPS}  \\
\hline
Dataset      &  CP Method & size $\downarrow$ &  coverage (\%)   &  size $\downarrow$ &  coverage (\%)         \\ \hline
TissueMNIST \cite{medmnistv2} &  CP & 2.35 $\pm$ 0.01  & 91.39 $\pm$ 0.19  & 2.33 $\pm$ 0.01 &  90.01 $\pm$ 0.23 \\
      (8 classes) &  Noisy-CP & 4.68 $\pm$ 0.04 &  99.34 $\pm$ 0.05 & 4.72 $\pm$ 0.04  & 99.27 $\pm$ 0.05 \\ 
                 &  NRES-CP & 4.27 $\pm$ 0.08 &  97.42 $\pm$ 0.12 & 4.40 $\pm$ 0.05  & 98.65 $\pm$ 0.13\\  
                  &  NR-CP & \textbf{3.27 $\pm$ 0.02} &  96.71 $\pm$ 0.08 & \textbf{3.40 $\pm$ 0.02}   & 96.94 $\pm$ 0.08\\ \hline 
PathMNIST   \cite{medmnistv2} &  CP & 2.74 $\pm$ 0.10 & 91.09 $\pm$ 0.54 & 2.71 $\pm$ 0.10   & 90.02 $\pm$ 0.70 \\
      (9 classes) &  Noisy-CP & 6.12 $\pm$ 0.14 &  96.36 $\pm$ 0.32 & 6.15 $\pm$ 0.13   & 96.43 $\pm$ 0.32\\
      &  NRES-CP & 4.47 $\pm$ 0.11 &  95.30 $\pm$ 0.21 & 6.02 $\pm$ 0.10  & 96.10 $\pm$ 0.15\\
                  &  NR-CP & \textbf{4.33 $\pm$ 0.08} &  94.01 $\pm$ 0.30 & \textbf{4.67 $\pm$ 0.08}   & 94.45 $\pm$ 0.29\\ \hline 
HAM10000  \cite{tschandl2018ham10000}   &  CP & 2.71 $\pm$ 0.27 &  90.68 $\pm$ 1.57 & 2.71 $\pm$ 0.27 & 90.21 $\pm$ 1.77 \\ 
      (7 classes) &  Noisy-CP & 4.98 $\pm$ 0.22 &  95.88 $\pm$ 0.96  & 4.99 $\pm$ 0.22   & 95.86 $\pm$ 0.94 \\ 
                 &  NRES-CP & 3.85 $\pm$ 0.17 &  94.52 $\pm$ 0.55 & 5.02 $\pm$ 0.15  & 96.33 $\pm$ 0.79 \\
                  &  NR-CP & \textbf{3.54 $\pm$ 0.12} &  92.77 $\pm$ 0.88 & \textbf{3.69 $\pm$ 0.15}   & 92.94 $\pm$ 0.90 \\ \hline 
OrganSMNIST \cite{medmnistv2} &  CP & 1.67 $\pm$ 0.02 &  93.33 $\pm$ 0.36 & 1.63 $\pm$ 0.02   & 90.02 $\pm$ 0.57 \\ 
      (11 classes) &  Noisy-CP & 6.15 $\pm$ 0.16 &  99.76 $\pm$ 0.05 &     6.23 $\pm$ 0.15  & 99.54 $\pm$ 0.08\\ 
                  &  NRES-CP & {5.96 $\pm$ 0.09} &  99.50 $\pm$ 0.16 & {6.31 $\pm$ 0.10}   & 99.61 $\pm$ 0.17 \\ 
                  &  NR-CP & \textbf{2.96 $\pm$ 0.06} &  98.35 $\pm$ 0.12 & \textbf{4.31 $\pm$ 0.22}   & 99.20 $\pm$ 0.11 \\ \hline 
\end{tabular}
    }
\end{table}

\comment{
\begin{figure}
    \centering
    \includegraphics[scale=0.46, , trim= 0 00 450 00 ,clip]{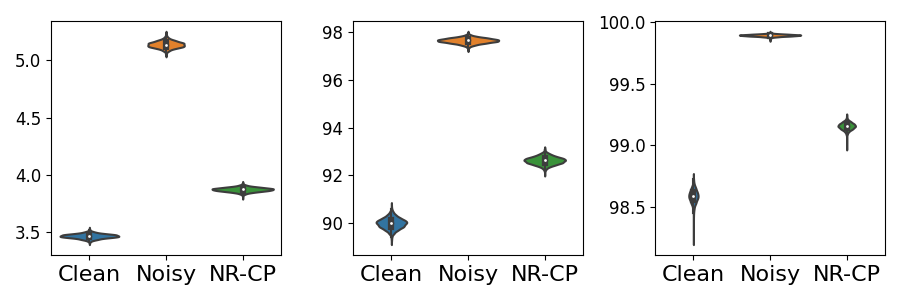}
     \includegraphics[scale=0.46, , trim= 200 00 220 00 ,clip]{images/average_aps/TissueMNIST/violin_new.png}
        \includegraphics[scale=0.27]{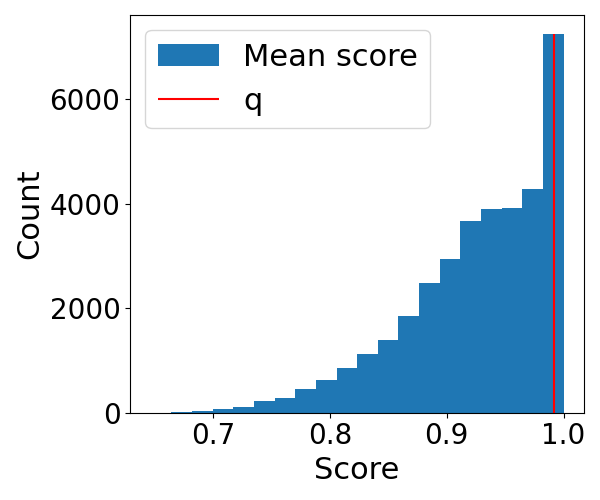}\\
      (a) \hspace{2.8cm}
    (b)   \hspace{2.8cm} (c)
      \caption{ Statistics computed on 5000 validation/test splits of the  TissueMNIST dataset. (a) Mean size and (b) coverage for APS, Noisy-APS and NR-APS  (c) Threshold $q_{\epsilon}$ and histogram of $S(x)$ for the  APS Score. }
       \label{stat}
\end{figure}
}

\comment{We showed above that the prediction set obtained by NR-CP is smaller than the one obtained by NRES-CP, when $S(x) \le q_{\epsilon}$. Fig. \ref{stat}c presents a histogram of $S(x)$ along with $q_{\epsilon}$. Results show that indeed for most $x$ $S(x) \le q_{\epsilon}$ holds, verifying the superiority of NR-CP over NRES-CP.}

\begin{figure}
    \centering
    \includegraphics[scale=0.48]{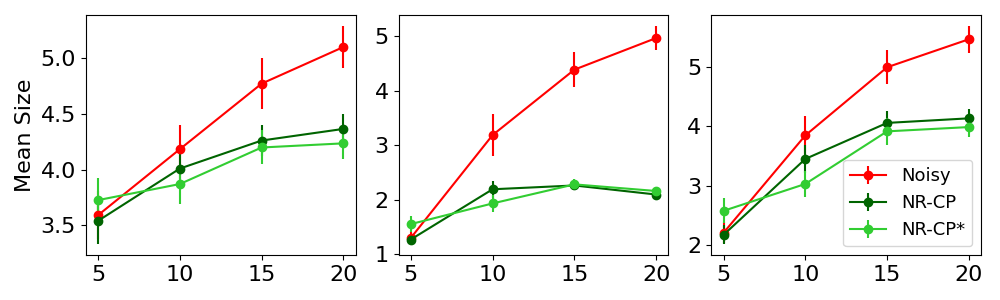}\\
    \includegraphics[scale=0.48]{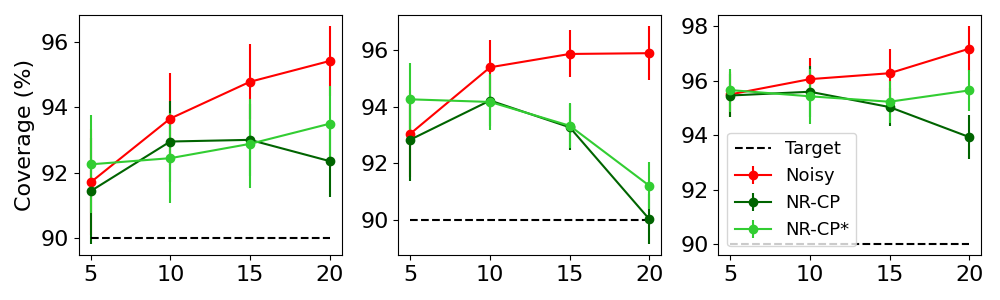}
    \hspace{3cm}
    (a) APS \hspace{2.5cm} (b) HPS \hspace{2.5cm}  (c) RAPS
    \caption{Mean size and coverage as a function of the noise level across different conformal prediction methods for HAM-10000 and $1\!-\!\alpha=0.9$.}
    \label{fig:epsilon_and_est_sweep}
\end{figure}

\textbf{CP Results.} Tables \ref{table_aps} and \ref{raps} report the noisy label calibration results for APS, and RAPS. 
 In all cases, we used $1\!-\!\alpha=0.9$ and a noise level of $\epsilon=.2$.   We report the mean and standard deviation over 5000 random splits.  Table 2 shows that in the case of a validation set with noisy labels, the CP threshold becomes larger to facilitate the uncertainty induced by the noisy labels. This yielded larger prediction sets and the coverage was higher than the target coverage which was set to $90\%$. The NR-CP method, which was aware of the noise rate, yielded better results in terms of the prediction set size, and approached the noise-free scenario. Both Noisy-CP and NRES-CP were under-confident and failed to reflect the uncertainty on the test set.

\textbf{End-to-end CP with noise level estimation.}
We next combined our NR-CP with a state-of-the-art noise-robust network training that estimated the noise level $\epsilon$ as part of the training \cite{li2021provably}.
Fig. \ref{fig:epsilon_and_est_sweep} reports the mean size and coverage results across different noise levels $\epsilon$ and different conformal prediction methods on the HAM-10000 dataset. We denote the NR-CP variant in which $\epsilon$ was estimated by NR-CP$^*$. The results show that as the noise rate increased, the noisy labels corrupted the calibration of the network more, as reflected in a larger mean size and higher coverage, whereas our method, even in the case where $\epsilon$ was estimated achieved much better calibration results. 

To conclude, we presented a procedure that makes it possible to apply the Conformal Prediction algorithm on a validation set with noisy labels. The main novelty of our approach is a transformation that renders a given
conformal score robust to label noise. We showed that our method outperforms current
methods by a large margin, in terms of the average size of the prediction
set, while maintaining the required coverage.
In this study, we concentrated on a simple noise model.  The proposed NR-CP procedure can  easily be extended to general noise models whose parameters can be extracted from the data by applying noise robust network learning methods.

\bibliographystyle{splncs04}
 \bibliography{refs}

\end{document}